\documentclass[letterpaper]{article} 
\usepackage[draft]{aaai2026}  
\usepackage{times}  
\usepackage{helvet}  
\usepackage{courier}  
\usepackage[hyphens]{url}  
\usepackage{graphicx} 
\usepackage{amsmath}
\usepackage{url}
\usepackage{amssymb}
\usepackage{booktabs}
\usepackage{amsthm}

\usepackage{tikz}
\usepackage{xcolor}
\usepackage{float}
\usepackage{tikz}
\usetikzlibrary{calc,spy}
\usepackage[font=small,skip=1pt]{caption}

\usepackage{subcaption}
\usetikzlibrary{shapes.geometric, arrows.meta, positioning}
\usepackage{natbib}  
\usepackage{caption} 
\usepackage{algorithm}
\usepackage{algorithmic}

\usepackage{newfloat}
\usepackage{listings}
\DeclareCaptionStyle{ruled}{labelfont=normalfont,labelsep=colon,strut=off} 
\floatstyle{ruled}
\newfloat{listing}{tb}{lst}{}
\floatname{listing}{Listing}
\title{PanoLora: Bridging Perspective and Panoramic Video Generation with LoRA Adaptation}
\author{
\begin{tabular}{c}
Zeyu Dong\thanks{Equal contribution. $^\dagger$ Corresponding authors.}$^{1}$,
Yuyang Yin\footnotemark[1]$^{2}$, 
Yuqi Li$^3$,
Eric Li$^{4}$, 
Hao-Xiang Guo$^{\dagger4}$, 
Yikai Wang$^{\dagger1}$
\end{tabular} \\[0.5em]   
\normalfont
\textsuperscript{1} School of Artificial Intelligence, Beijing Normal University \\ 
\textsuperscript{2}  Beijing Jiaotong University 
\textsuperscript{3} The City College of New York
\textsuperscript{4} Skywork AI
}

\usepackage{bibentry}

\usepackage{float}
\begin{document}

\maketitle

\begin{abstract}
Generating high-quality 360° panoramic videos remains a significant challenge due to the fundamental differences between panoramic and traditional perspective-view projections. While perspective videos rely on a single viewpoint with a limited field of view, panoramic content requires rendering the full surrounding environment, making it difficult for standard video generation models to adapt. Existing solutions often introduce complex architectures or large-scale training, leading to inefficiency and suboptimal results.
Motivated by the success of Low-Rank Adaptation (LoRA) in style transfer tasks, we propose treating panoramic video generation as an adaptation problem from perspective views. Through theoretical analysis, we demonstrate that LoRA can effectively model the transformation between these projections when its rank exceeds the degrees of freedom in the task. Our approach efficiently fine-tunes a pretrained video diffusion model using only approximately 1,000 videos while achieving high-quality panoramic generation. Experimental results demonstrate that our method maintains proper projection geometry and surpasses previous state-of-the-art approaches in visual quality, left-right consistency, and motion diversity. Project page: \textcolor{blue}{\linebreak\url{https://anonymous-pano-lora.github.io}}.

\end{abstract}

\section{Introduction}

Traditional perspective videos, the dominant format in media, capture scenes from a single viewpoint with a limited field of view (FoV), projecting 3D scenes onto a 2D plane. While suitable for conventional applications like films and online content, this format lacks spatial immersion. In contrast, panoramic videos expand the FoV to 360° horizontally and 180° vertically using specialized cameras or multi-camera arrays~\cite{miller1996panorama, gledhill2003panoramic, li2004stereo, ferreira2007panorama, gao2022review}. By projecting onto spherical or cylindrical surfaces, they enable users to freely explore the environment by rotating their viewpoint—delivering a more immersive and realistic experience. This makes them particularly valuable for VR, storytelling, and simulations~\cite{jayaram2001assessment, slater2016enhancing, radianti2020systematic, conrad2024learning}.

However, generating high-quality panoramic videos remains challenging. Traditional video generation models~\cite{wan2025, yang2024cogvideox, sun2024hunyuan, wang2024svd}, trained primarily on perspective-view data, struggle to adapt to panoramic camera motions due to fundamental differences in projection. Unlike perspective videos, which rely on a single viewpoint, panoramic videos require rendering the entire surrounding environment, posing unique technical hurdles.

Previous methods for panoramic video generation~\cite{wang2024360dvd,lu2024genex,tan2024imagine360,ye2024diffpano,liu2025dynamicscaler} primarily address narrow fields by introducing additional model architectures to achieve perspective-to-panoramic transformation. For instance, these approaches often employ large-parameter encoders to generate control signals for panoramic images or auxiliary network blocks to inject supplementary information into the main network.
However, these solutions typically require extensive training data to optimize their substantial parameter sizes and often yield suboptimal results. 
On the other hand, Low-Rank Adaptation (LoRA)~\cite{hu2022lora} has demonstrated significant effectiveness in fine-tuning generative models, particularly for generating content with specific styles.
This leads us to pose the following question: \textit{Can we treat panoramic video generation as a style transfer task from perspective views and accomplish it efficiently using lightweight LoRA adaptation?}

To validate this idea, we first analyze the transformation process from the perspective view to panoramic view images. The pipeline involves projecting perspective images onto 3D points using camera intrinsic and extrinsic parameters, followed by an equirectangular projection onto the panoramic plane. Based on the characteristics of the dataset, we identify that this transformation involves at least eight degrees of freedom.
Next, we investigate whether LoRA can effectively model this transformation from the perspective of solution space. Through rigorous theoretical analysis, we examine the relationship between model output variations from fine-tuning, the rank of LoRA, and the solution space. Our findings demonstrate that, for both single-layer and multi-layer networks, LoRA can achieve the desired transformation as long as its rank exceeds the degrees of freedom in the task.

We conduct extensive experiments and find that with only a small dataset (approximately 1,000 videos), we can efficiently fine-tune a pretrained video diffusion model for panoramic video generation. The results show that when the rank value is below 8, the limited solution space causes distorted outputs that violate projection rules. In contrast, higher rank values demonstrate superior performance. Furthermore, since we verify that sufficient solution space exists for panoramic generation tasks, our method achieves significant improvements over previous approaches in terms of visual quality, left-right consistency and motion diversity.

In summary, our contributions are threefold: (1) we reformulate panoramic video generation as a task suited for LoRA-based adaptation, (2) we theoretically demonstrate that LoRA can effectively adapt pretrained video diffusion models to panoramic video generation, and (3) our model achieves state-of-the-art performance both qualitatively and quantitatively. These results highlight LoRA's unique capability to bridge the gap between standard and panoramic views, enabling more immersive and interactive 3D scene generation.

\section{Related Works}
\subsection{Panoramic Video Generation}
With the growing demand for immersive VR/AR content, panoramic generation has progressed from still images to dynamic videos.
360DVD~\cite{wang2024360dvd} is the first diffusion-based framework explicitly tailored for controllable panoramic video via spatial conditioning on fine-tuned $360^\circ$ data; while it yields globally consistent frames, its outputs are largely static with weak temporal dynamics. Subsequent work like PanoDream~\cite{paliwal2024panodreamer} adds camera-motion synthesis but often relies on fixed motion priors or scene templates, limiting adaptability.
DynamicScaler~\cite{liu2025dynamicscaler} extends to long-loop panoramic \emph{videos} with improved frame consistency, yet remains limited in extensibility.
SpotDiffusion~\cite{frolov2025spotdiffusion} and VidPano~\cite{ma2024vidpanos} explore prompt-driven and efficient panoramic video synthesis but still struggle to control complex, long-range scene dynamics. PanoVerse~\cite{pintore2023panoverse} unifies $360^\circ$ image/video/depth generation, at the cost of heavy training and limited semantic control. TiP4GEN~\cite{xing2025tip4gen} targets text-to-dynamic panoramic scenes with fine-grained content control and geometry-consistent 4D reconstruction.

We focus on the fundamental challenges of the text-to-panoramic video task. Our method introduces a lightweight low-rank adaptation (LoRA) across spatial and temporal layers to learn continuous camera trajectories and controllable dynamics, delivering smoother motion and stronger temporal coherence while preserving panoramic seam closure.

Concurrent work, UniPano~\cite{ni2025makes}, explores text-to-360° panoramic image generation by analyzing the roles of components such as query, key, and value matrices in the generation process. In contrast, our work primarily addresses the text-to-panoramic video generation task, which introduces new challenges, such as maintaining both spatial and temporal consistency with motion. We also analyze why LoRA is effective in adapting a perspective video generation base model to panoramic video generation.


\subsection{3D Scene and Camera Generation}
Recent 3D-aware generative models, for example, 3D-SceneDreamer~\cite{zhang20243d}, 4REAL~\cite{yu20244real}, DimensionX~\cite{sun2024dimensionx}, Diffusion4D~\cite{liang2024diffusion4d} and other works~\cite{ren2025gen3c,yin20234dgen} enable dynamic 3D scene synthesis and text-driven camera trajectories. However, they often target object-centric or small-scale scenes and lack support for full panoramic generation or LoRA-based modular control. Our method fills this gap by focusing on full-scene panoramic dynamics with efficient adaptation. Additionally, while 3D-aware models typically require volumetric rendering or scene reconstruction, our approach operates directly on 2D panoramic projections with minimal computational overhead. Through LoRA, we preserve the benefits of pretrained video diffusion models while achieving 3D-like motion realism across the full panoramic sphere.



\begin{figure*}
    \begin{center}\vspace{-0.2cm}
        \includegraphics[width=0.95\linewidth]
        {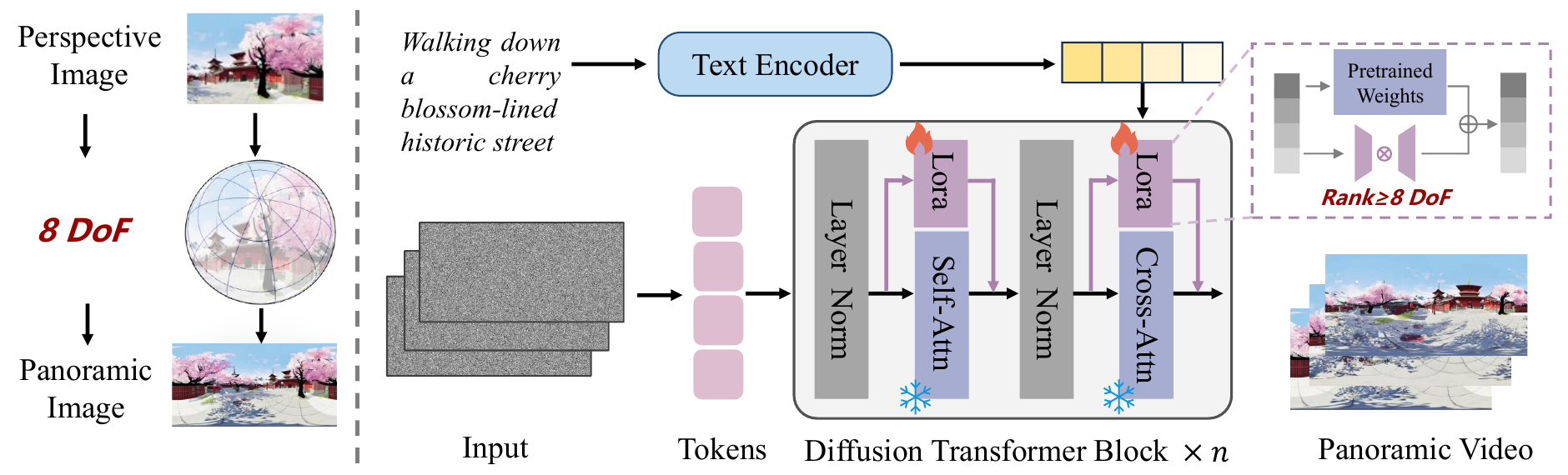}
        \vspace{-1mm}
        \caption{Our goal is to generate high-quality panoramic videos using pretrained video diffusion models. We first analyze how to transform perspective images into panoramic form and demonstrate that this transformation requires at least eight degrees of freedom. Subsequently, we employ LoRA (Low-Rank Adaptation) for efficient fine-tuning and provide theoretical proof that LoRA can adequately cover this transformation solution space.}
        \label{fig:method}
    \end{center}
    \vspace{-0.3cm}
\end{figure*}

\section{Method}
\subsection{Task Definition}
Our goal is to use a pre-trained video diffusion generation model to generate panoramic video. 
We start by analyzing the degrees of freedom (dof) involved in the transformation between a perspective view image $\mathbf{x}_p$ and its corresponding panoramic view image $\mathbf{x}_{360} $. The process involves three main steps: (1) \textbf{Restoring camera coordinates:} Since the perspective view is derived from the pinhole camera model, the image coordinate system can be restored to the camera coordinate system using intrinsic parameters. (2) \textbf{Transforming to world coordinates:}  The points in the camera coordinate system are then transformed into the world coordinate system via extrinsic parameters, converting them into 3D spatial points. (3) \textbf{Projecting 3D points onto a panoramic plane:} These 3D points are mapped onto a unit sphere and projected onto a plane using equirectangular projection, resulting in the corresponding panoramic image. The complete transformation from a perspective image to a panoramic image can be expressed as
\begin{equation}
\mathbf{x}_{360} = \Psi(K, R, t; \mathbf{x}_p),
\end{equation}
where $K \in \mathbb{R}^{3\times3}$ is the camera intrinsic matrix with 5 dof, which includes the focal lengths $f_x, f_y$, principal point offsets $c_x, c_y$, and the skew coefficient $s$; $R \in SO(3)$ is the rotation matrix with {3 dof}; and $t \in \mathbb{R}^3$ is the translation vector with {3 dof}. Since the equirectangular projection does not require additional parameters, the total degrees of freedom for the projection transform are
\begin{equation}
\dim(\mathfrak{g}_p) = 5\,(\text{intrinsic}) + 6\,(\text{SE(3)}) = 11.
\end{equation}



However, an 11-degree-of-freedom (DoF) transformation remains overly complex. We aim to simplify this problem through careful analysis of our target dataset.
We analyze the camera pose distribution in 447 full-scene video clips from the training dataset, covering urban and natural scenes. These clips range from 2 to 6 seconds in duration, with a frame rate of 24 FPS, and a resolution of 768$\times$768 pixels.  The statistical results are shown in Table~\ref{tab:camera_stats}. The standard deviations of the horizontal shift ($t_x$) and forward shift ($t_y$) are 833.11 $\mathrm{m}$ and 3066.65 $\mathrm{m}$, respectively. The standard deviation of yaw angle ($\theta_{\text{yaw}}$) is 108.23°, indicating substantial viewpoint variation in full scenes. These parameters—$t_x$, $t_y$, and $\theta_{\text{yaw}}$—exhibit higher variability. In contrast, the shift $t_z$ is 7.92 $\mathrm{m}$, with pitch 1.05° and roll 1.18°, respectively. These parameters are relatively more stable.

\begin{table}[h]
    \centering
    \caption{Camera Motion Parameter Statistics.}
    \label{tab:camera_stats}
    \begin{tabular}{lcc}
        \toprule
        Parameter & Std. Deviation & Mean \\
        \midrule
        Horizontal shift ($t_x$) & 833.11 $\mathrm{m}$ & 0.15 $\mathrm{m}$ \\
        Forward shift ($t_y$) & 3066.65 $\mathrm{m}$ & 2.34 $\mathrm{m}$ \\
        Vertical shift ($t_z$) & 7.92 $\mathrm{m}$ & 0.02 $\mathrm{m}$ \\
        Pitch($\theta_{\text{pitch}}$) & 1.05° & 0.03° \\
        Yaw($\theta_{\text{yaw}}$) & 108.23° & 0.12° \\
        Roll($\theta_{\text{roll}}$) & 1.18° & 0.01° \\
        \bottomrule
    \end{tabular}
\end{table} 
Based on the above analysis, we empirically simplify the camera motion model, retaining only the horizontal shift ($t_x$), forward shift ($t_y$), yaw angle ($\theta_{\text{yaw}}$), focal lengths $f_x$, $f_y$, optical center $c_x$, $c_y$, and scale factor $s$, resulting in an 8-dof model. The panoramic projection function is defined as:
\begin{equation}
    \mathbf{x}_{360} = \Psi(f_x, f_y, c_x, c_y, s, t_x, t_y, \theta_{\text{yaw}}; \mathbf{x}_p).
\label{eq:8dof}
\end{equation}
In total, the transformation from a perspective image to a panoramic image can be summarized as an 8-degree-of-freedom (DoF) problem.


\subsection{Efficient Model Fine-Tuning}
Previous conditional video generation~\cite{wang2024360dvd,bahmani2025ac3d,liang2025wonderland} handles these 8-DoF controls by adding an additional branch (\emph{e.g.}, a camera control module) to the DiT framework. This module guides the attention mechanism, enabling consistent 360° video synthesis. However, this approach faces key limitations: it requires training auxiliary networks with substantial parameters to fine-tune the primary model and depends on substantial training data to achieve satisfactory performance. These factors make the task particularly challenging. 

An alternative video diffusion fine-tuning strategy avoids introducing new network architectures and instead focuses on minimal parameter adjustments to the original branch network. The most common implementation of this approach uses Low-Rank Adaptation (LoRA). Unlike full fine-tuning, LoRA employs two low-rank matrices to approximate weight updates: (1) a down-projection matrix $A$ that reduces the input dimension, (2)an up-projection matrix $B$ that restores the original dimension.
These matrices collectively represent $\Delta W=BA$, where the product BA forms a low-rank approximation of the weight update matrix. This decomposition dramatically reduces the number of trainable parameters while maintaining effective adaptation capability. 

LoRA is typically employed for customized style generation tasks rather than conditional generation. However, although our task (transforming perspective videos to panoramic videos) requires eight-degree-of-freedom parameters as conditional signals, it can fundamentally be viewed as a style transformation task—converting perspective-style content into panoramic-style output.

Given these analyses, we ask: \textit{Can LoRA—the simplest fine-tuning approach—effectively accomplish this task?} Specifically, we fine-tune the DiT block in state-of-the-art video generation models (\emph{e.g.}, Wan2.1~\cite{wan2025}) using LoRA, formulated as
$$\mathbf{h}' = \mathbf{h} + \alpha \cdot \mathbf{W}_{A}\mathbf{W}_{B}\mathbf{h},$$
where $\mathbf{h}$ is the original hidden state.
$\mathbf{W}_{A} \in \mathbb{R}^{d \times r}$ and $\mathbf{W}_{B} \in \mathbb{R}^{r \times d}$ are the low-rank adaptation matrices.
$r \ll d$ is the rank of the adaptation and 
$\alpha$ is a scaling factor controlling the adaptation strength.

To rigorously validate the validity of our hypothesis, we provide a formal proof below.



\subsection{Theoretical Proof}
We aim to transform a perspective video generation model into a panoramic video generation model. Based on previous work, this transformation requires a change of 8 degrees of freedom (DoF). Specifically, the goal is to prove that a LoRA modification can cover the required solution space.

\noindent\textbf{Definition of output change.} Let the model be represented by a function \( f_\theta: \mathbb{R}^p \to \mathcal{Y} \), where \( \theta \in \mathbb{R}^p \) represents the model parameters, and \( \mathcal{Y} \) denotes the output video space. For a fixed input \( x \), define the function \( F(\theta) := f_\theta(x) \). The output space \( \mathcal{Y} \in \mathbb{R}^{F\times H \times W \times C} \), where \(F\, H \), \( W \), and \( C \) denote the frames, height, width, and number of channels, respectively. Applying the first-order Taylor expansion around \( \theta \), we approximate the output variation as
$$
F(\theta + \Delta\theta) \approx F(\theta) + J_F(\theta) \cdot \Delta\theta,
$$
where \( J_F(\theta) \in \mathbb{R}^{q \times p} \) is the Jacobian of the function \( f_\theta(x) \) with respect to \( \theta \), and \( \delta F := J_F(\theta) \cdot \Delta\theta \) denotes the change in output. Physically, \( \delta F \) represents how the output video varies in response to a small perturbation \( \Delta\theta \) in the model parameters.

\noindent\textbf{Low-rank adaptation for the single-layer case.} In LoRA, the parameter change \( \Delta\theta \) is constrained to lie in a low-rank subspace
$$
\Delta\theta = AB, \quad A \in \mathbb{R}^{p \times r}, B \in \mathbb{R}^{r \times 1}.
$$

The rank of the output variation is determined by the Jacobian $J_F(\theta) $ and low-rank structure of $\Delta\theta$ by
$$
\operatorname{rank}(\delta F) =\operatorname{rank}(J_F(\theta) \cdot \Delta\theta ).
$$

By the properties of matrix multiplication, specifically the inequality \( \operatorname{rank}(XY) \leq \min(\operatorname{rank}(X), \operatorname{rank}(Y)) \), we obtain the following rank condition for the output change
$$
\operatorname{rank}(\delta F) \leq \min(\operatorname{rank}(J_F), \operatorname{rank}(\Delta\theta)) \leq r.
$$

For the transformation to span a subspace of dimension \( d \), the rank \( r \) must satisfy \( r \geq d \). Since the transformation from a perspective to a panoramic view requires 8 degrees of freedom, we conclude that if the LoRA rank \( r \geq 8 \), the changeable space can be fully covered. If the network is a single layer, this is sufficient to span the required space.

\noindent\textbf{Multi-layer linear network case.}  
We now consider the special case where the mulit-layer network is composed entirely of linear transformations. Specifically, let the model be represented as
$$
f_\theta = f_L \circ f_{L-1} \circ \dots \circ f_1,
$$
where each \( f_l \) is a linear layer of the form
\[
f_l(x) = A_l x + b_l,
\]
which is commonly used in deep neural networks. Here, \( A_l \in \mathbb{R}^{n_{l+1} \times n_l} \) is the weight matrix, and \( b_l \in \mathbb{R}^{n_{l+1}} \) is the bias term. The total parameter set is \( \theta = (\theta_1, \dots, \theta_L) \), where \( \theta_l = (A_l, b_l) \).

When LoRA modules \( \Delta\theta_l = A_l B_l \) of rank \( r \) are inserted into each layer, the first-order output change becomes
\[
\delta F = \sum_{l=1}^L J_F^{(l)}(\theta) \cdot \Delta\theta_l.
\]

In this linear case, each Jacobian \( J_F^{(l)}(\theta) \) corresponds directly to the weight matrix of the layer (up to input-dependent coefficients), and the output change can be expressed as a sum of low-rank contributions from each layer. If each \( \Delta\theta_l \) has rank at least 8, then the total variation \( \delta F \) will span a space of dimension at least 8. We thus obtain a relatively loose but sufficient lower bound: the sum of per-layer low-rank changes yields an overall rank capable of covering the 8 degrees of freedom required for the transformation.

\noindent\textbf{Extension to the non-linear networks.} For non-linear networks, such as those incorporating attention mechanisms in video generation models, the Jacobian \( J_F^{(l)}(\theta) \) remains well-defined, and the activation functions(like ReLU, Softmax, etc) are locally differentiable. Since the local behavior of these functions is approximately linear in small neighborhoods, the rank condition from the linear case extends to non-linear layers as well. Therefore, as long as each layer has a LoRA rank \( r \geq 8 \), the total rank of \( \delta F \) will be sufficient to cover the 8 degrees of freedom required for the transformation.

\noindent\textbf{Conclusion.} We have shown that for both single-layer and multi-layer network cases, applying LoRA with a rank \( r \geq 8 \) allows the model to capture the 8 degrees of freedom required for the transformation from perspective video generation to panoramic video generation. This result holds for both linear and modern non-linear architectures, completing the proof.

\section{Experiments}
\subsection{Implement Details}
To achieve high-quality panoramic video generation with semantic alignment and dynamic consistency, we built a structured pipeline centered around a custom training dataset. Specifically, we collected panoramic video sequences using a UE-based environment, enabling precise control over camera motion and scene content. These videos serve as the training data for our model, supporting supervised learning of realistic dynamics and spatial consistency. Next, Qwen-VL 2.5~\cite{team2024qwen2} is leveraged to interpret the visual content and generate semantically rich prompts, which guide the subsequent video synthesis process.
 
These prompts are then used to fine-tune the Wan2.1-14B video generation model via the PEFT~\cite{peft} toolkit: specifically, LoRA modules are injected into and trained on key layers of the model, including the query, key, value, and output (QKVO) projections in self-attention layers, cross-attention layers, and feed-forward networks (FFNs). This targeted fine-tuning strategy aims to endow ordinary video generation models with the capability of panoramic video synthesis.
 
To further optimize the performance, we conducted comparative experiments on different configurations: we tested various combinations of LoRA injection positions (\emph{e.g.}, standalone self-attention layers, combined cross-attention and FFN layers) and evaluated the impact of different LoRA ranks (5, 8, 16, 32) on the generated results.
 
Using the fine-tuned models, we generate panoramic videos and extract representative sequences through uniform frame sampling for downstream evaluation. The model is trained on a curated dataset of 1,000 panoramic videos over 15 epochs with a batch size of 1. Training is conducted on 8 NVIDIA A100 GPUs and takes approximately 10 hours to complete. This setup ensures that the model effectively captures both spatial coherence and temporally consistent dynamics for high-quality 360° video generation.

\subsection{Evaluation Metrics}
To rigorously evaluate the quality of generated panoramic videos, we adopt two key metrics that reflect the unique spatial and temporal characteristics of 360° content: \textbf{Left–right consistency} and \textbf{Inter-frame motion magnitude}. Since existing evaluation pipelines focus primarily on image quality or global coherence, they overlook critical aspects such as seam closure and motion strength across viewpoints. We therefore introduce tailored metrics to capture these factors.

\paragraph{Left-right consistency (L-R).}
This metric evaluates the spatial continuity at the seam connecting the leftmost and rightmost edges of each panoramic frame, corresponding to the $-180^\circ$ and $+180^\circ$ boundaries. In an ideal 360° projection, these edges should align perfectly to form a seamless visual loop.
As no prior work explicitly measures this form of consistency, we propose a tailored evaluation method. For each frame, we extract narrow vertical strips (2 pixels wide) from both the left and right boundaries. These strips are flattened into vectors of size $H \times 2 \times 3$ (height × width × RGB channels), capturing the color patterns along the seam. We then compute the cosine similarity between the two vectors to quantify alignment—higher values indicate better consistency, with a score of 1.00 representing perfect closure.
This metric is sensitive to discontinuities or visual flickering at the seam and provides a reliable signal for detecting projection artifacts that would otherwise degrade the immersive experience of panoramic videos.

\paragraph{Inter-frame motion magnitude.}
We compute the average dense optical flow magnitude (using the Farneback algorithm~\cite{farneback2003two}) between consecutive frames across four canonical directions: front, back, left, and right. This metric reflects the strength of motion and serves as a proxy for camera movement or dynamic scene changes. Higher values indicate more immersive temporal behavior.

\begin{itemize}
    \item Low motion magnitude implies static and less realistic content.
    \item High motion magnitude suggests active viewpoint transitions and richer dynamics.
\end{itemize}
\begin{figure*}[t]
\centering
\makebox[\textwidth][c]{
  \begin{minipage}[t]{0.19\textwidth}
    \includegraphics[width=\linewidth,trim=4 4 4 4,clip]{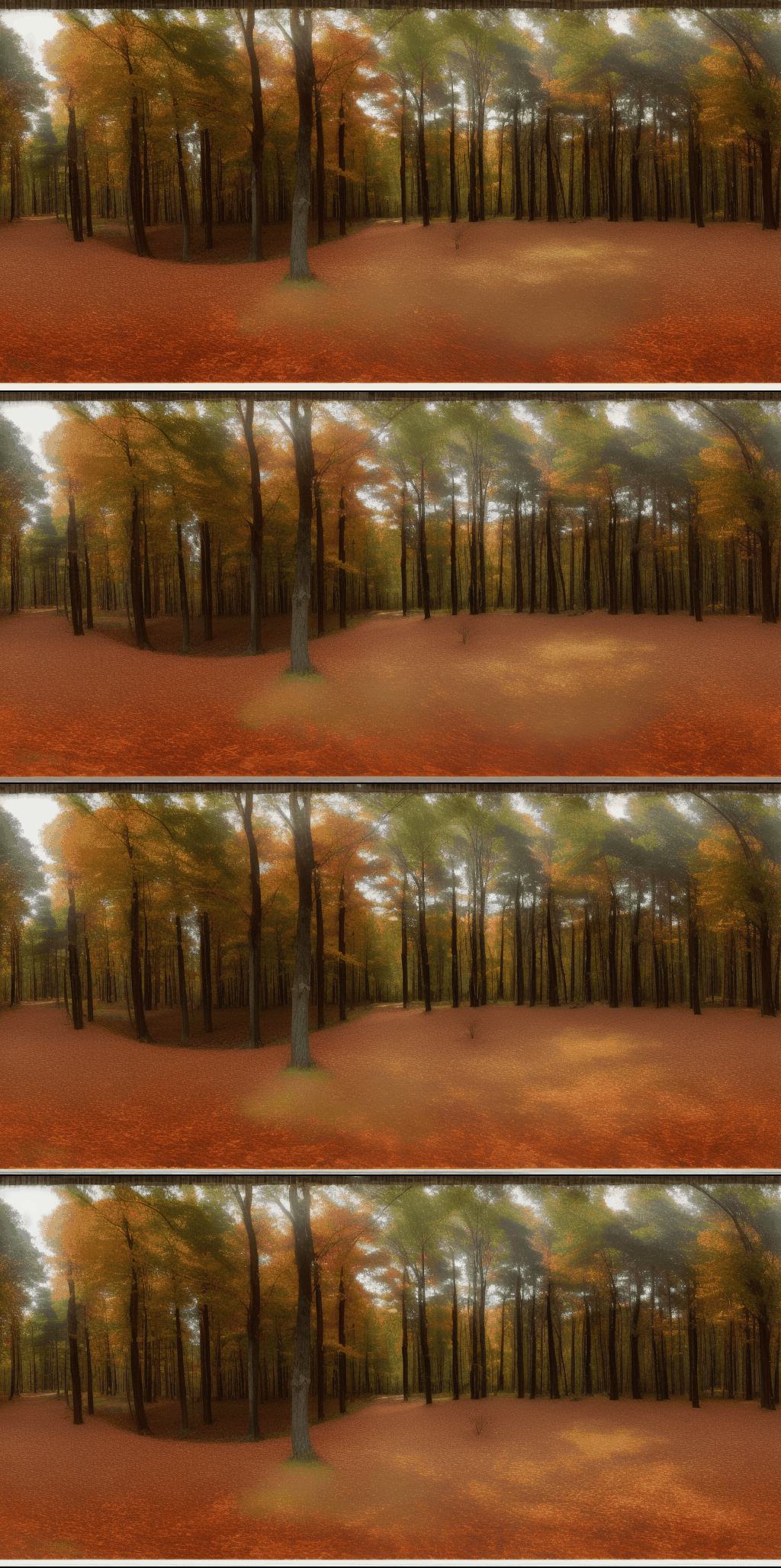}
    \vspace{2pt}\centering\scriptsize (a) 360DVD
  \end{minipage}\hspace{4pt}%
  \begin{minipage}[t]{0.19\textwidth}
    \includegraphics[width=\linewidth,trim=4 4 4 4,clip]{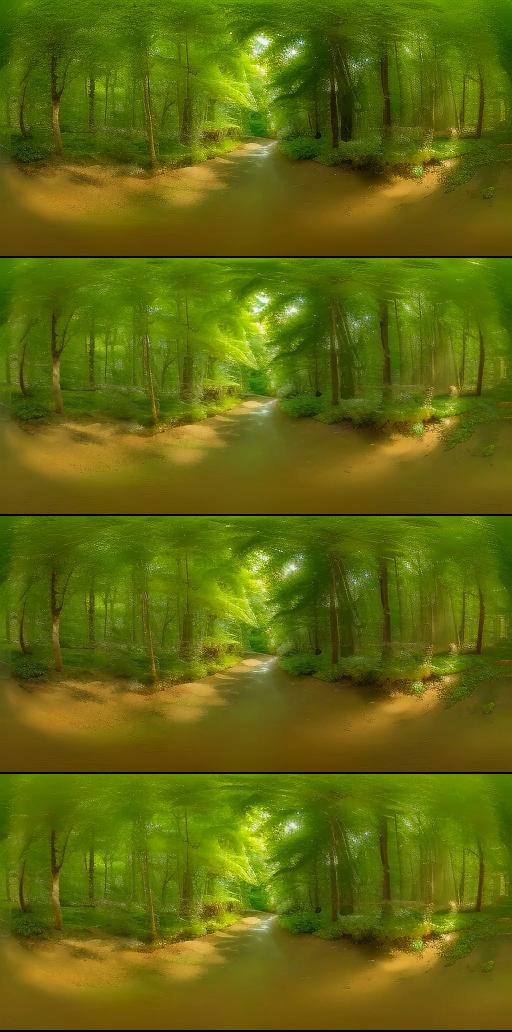}
    \vspace{2pt}\centering\scriptsize (b) 360DVD$^+$
  \end{minipage}\hspace{4pt}%
  \begin{minipage}[t]{0.19\textwidth}
    \includegraphics[width=\linewidth,trim=4 4 4 4,clip]{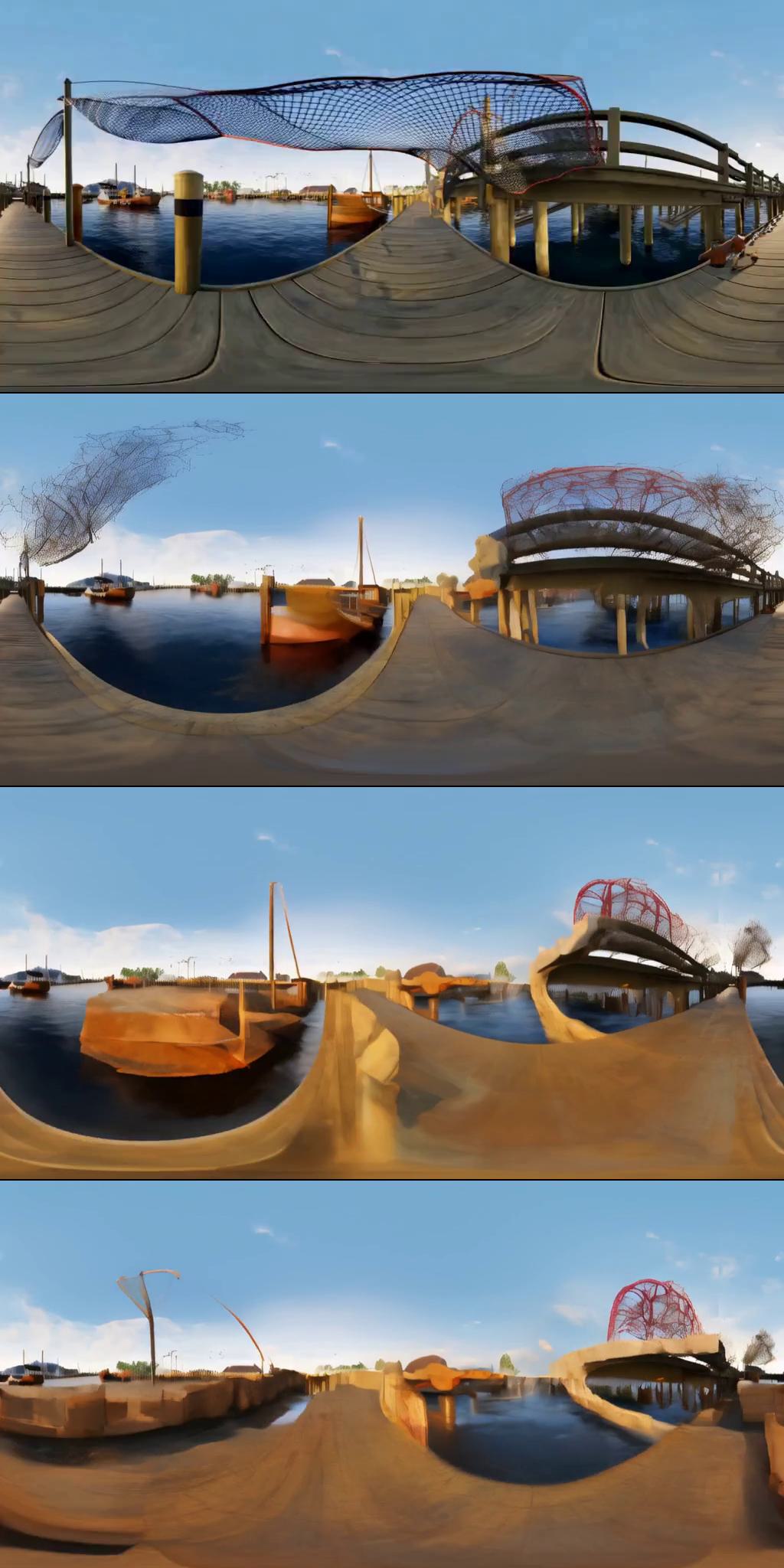}
    \vspace{2pt}\centering\scriptsize (c) GenEX
  \end{minipage}\hspace{4pt}%
  \begin{minipage}[t]{0.19\textwidth}
    \includegraphics[width=\linewidth,trim=4 4 4 4,clip]{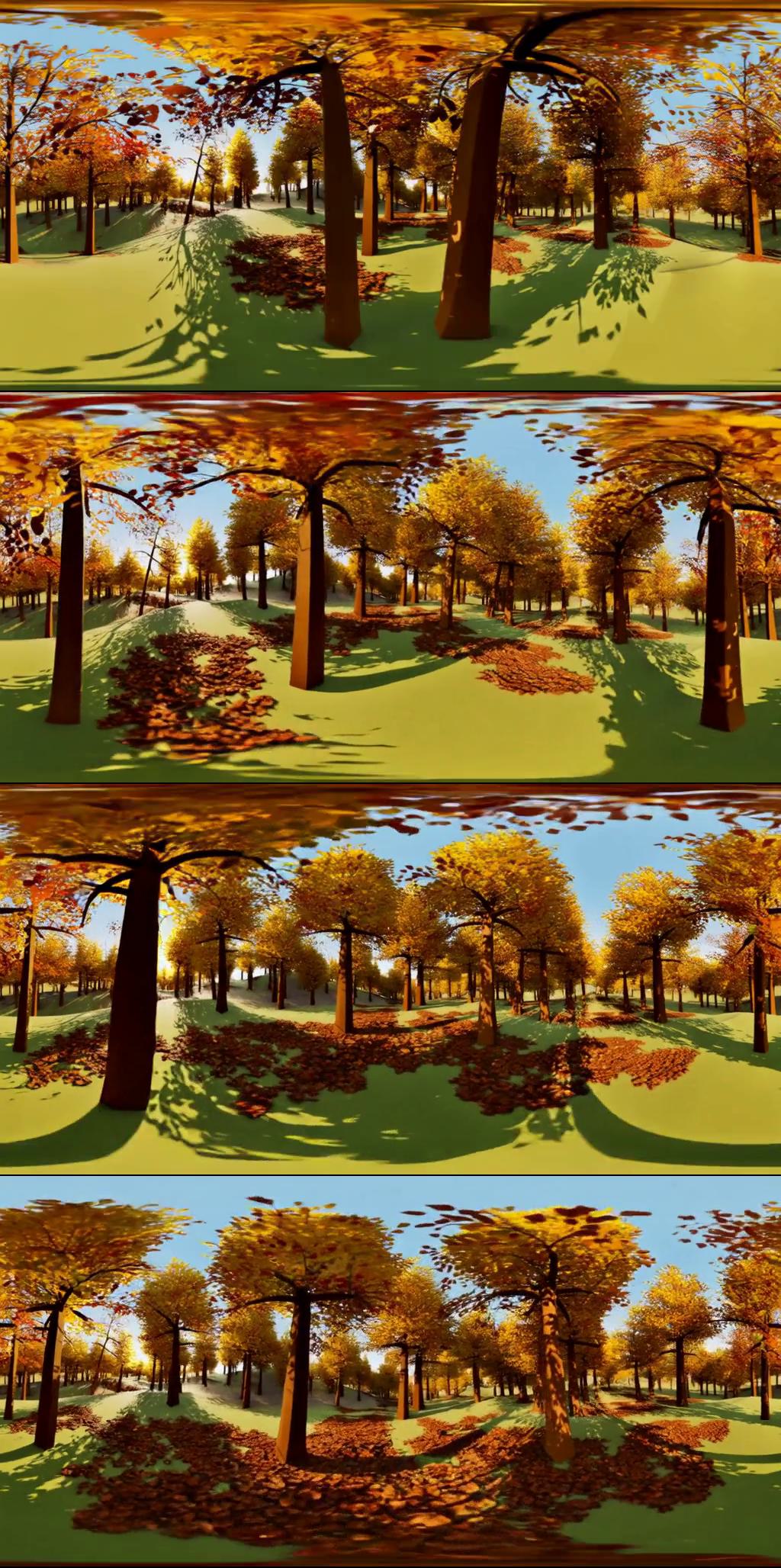}
    \vspace{2pt}\centering\scriptsize (d) Ours
  \end{minipage}\hspace{4pt}%
  \begin{minipage}[t]{0.19\textwidth}
    \includegraphics[width=\linewidth,trim=4 4 4 4,clip]{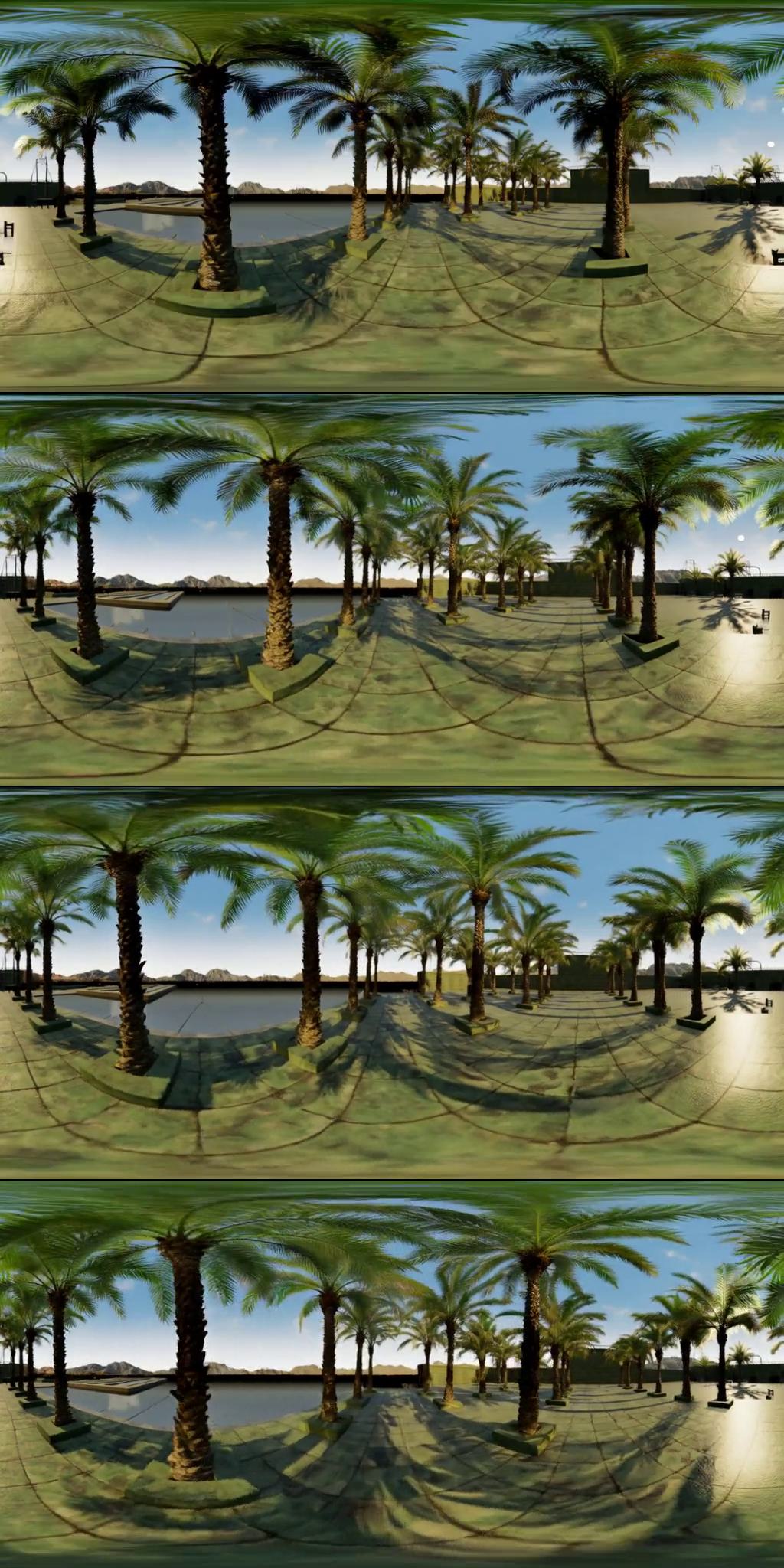}
    \vspace{2pt}\centering\scriptsize (e) Ours
  \end{minipage}%
}%
\caption{\textbf{Qualitative comparison across baselines.} Each column shows four consecutive equirectangular frames.
\textbf{360DVD}: largely static; \textbf{360DVD$^+$}: limited dynamics with faint seam blur; \textbf{GenEX}: temporal degradation (distortions/blur);
\textbf{Ours}: sharper details and smoother temporal transitions. View in color and zoom for detail.}
\label{fig:frames_compare}
\end{figure*}

\subsection{Main Results}
\begin{figure}[h]
\centering
\begin{subfigure}[b]{\linewidth}
  \centering
  \includegraphics[width=\linewidth]{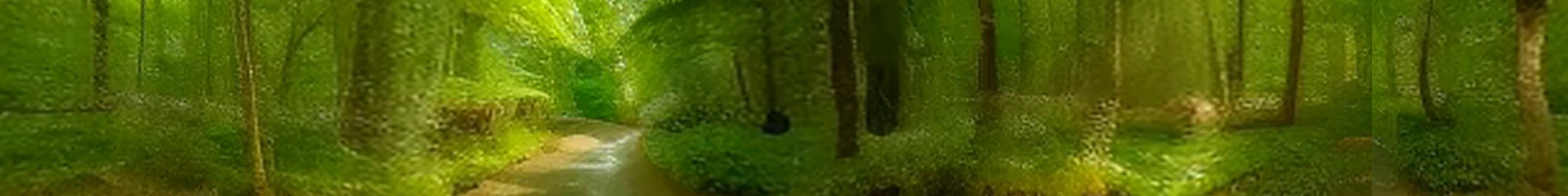}
  \caption{360DVD$^+$. Foreground remains nearly static; faint blur bands indicate seam misalignment.}
  \label{fig:persp_360}
\end{subfigure}
\begin{subfigure}[b]{\linewidth}
  \centering
  \includegraphics[width=\linewidth]{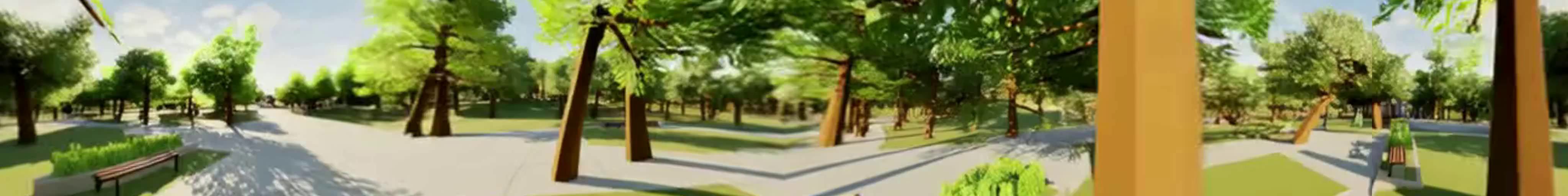}
  \caption{Ours. Continuous parallax and clean seams yield immersive, geometrically consistent motion.}
  \label{fig:persp_ours}
\end{subfigure}
\caption{Perspective comparison of four viewpoints concatenated into a single strip. Our model preserves spatial continuity and realistic camera dynamics better than 360DVD$^+$.}
\label{fig:perspective}
\end{figure}

Given the limited prior work on text-to-panoramic-video generation, we benchmark our approach against 360DVD~\cite{wang2024360dvd} using both quantitative and qualitative evaluations. To assess the generality of our LoRA-based adaptation, we extend the original 360DVD by injecting low-rank modules into its spatial and temporal layers, yielding \textbf{360DVD$^+$} (Figure~\ref{fig:frames_compare} (b)). Unlike the original 360DVD (Figure~\ref{fig:frames_compare} (a)), which predominantly produces static panoramas, 360DVD$^+$ introduces moderate dynamics with improved temporal coherence. We also compare to \textbf{GenEX}~\cite{lu2024genex}, a recent state-of-the-art general video diffusion framework. Since GenEX is not tailored for panoramic generation, we condition it on the first frame of our generated sequence and evaluate its ability to maintain coherent, realistic $360^\circ$ motion. Quantitative metrics are reported in Table~\ref{tab:metrics_360dvd}; qualitative results are provided in Figures~\ref{fig:frames_compare} and \ref{fig:perspective}.

\begin{table}[h]
\centering
\caption{Comparison among 360DVD, 360DVD$^+$, GenEX, and our method. Metrics include left–right consistency (L–R Cons.) ($\uparrow$) and motion magnitude across views (front, back, left, right) ($\uparrow$).}
\label{tab:metrics_360dvd}
\begin{tabular}{lccccc}
\toprule
\textbf{Model} & \textbf{L--R Cons.} & \textbf{Front} & \textbf{Back} & \textbf{Left} & \textbf{Right} \\
\midrule
GT             & 1.00 & 7.64 & 6.70 & 7.71 & 7.69 \\
360DVD         & 0.99 & 0.98 & 1.00 & 1.10 & 1.11 \\
360DVD$^+$     & 0.96 & 1.50 & 1.23 & 1.10 & 1.53 \\
GenEX          & 0.99 & 1.91 & 1.48 & 3.20 & 4.14 \\
Ours           & \textbf{0.99} & \textbf{4.02} & \textbf{3.99} & \textbf{3.56} & \textbf{5.11} \\
\bottomrule
\end{tabular}
\end{table}

\noindent\textbf{Quantitative comparison.}
Table~\ref{tab:metrics_360dvd} summarizes spatial and temporal quality. 
We measure \emph{motion magnitude} as the average frame-to-frame displacement (optical-flow norm) within each perspective crop; higher values indicate stronger apparent camera motion and depth-consistent parallax, whereas lower values imply near-static imagery.

Both 360DVD and our method (LoRA rank 16) achieve \textbf{0.99} L--R consistency, evidencing pixel-level $360^\circ$ seam closure. 
In contrast, 360DVD$^+$ drops to 0.96, showing residual misalignment. 
Moreover, lower-capacity (rank 5) and higher-capacity (rank 32) variants of our model\footnote{See Sec.\,4.4 for the ablation study of LoRA ranks.} either under-fit (visible discontinuities) or over-fit (unstable seams).

\begin{figure*}[t]
\centering
\captionsetup{skip=2pt}
\captionsetup[subfigure]{aboveskip=2pt, belowskip=0pt}

\begin{subfigure}[t]{0.49\linewidth}
  \centering
  \includegraphics[width=\linewidth]{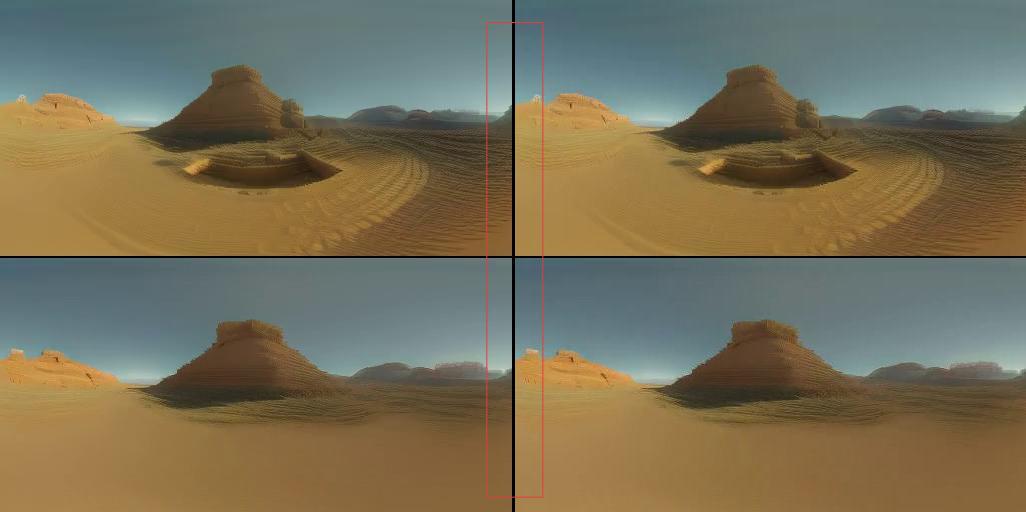}
  \caption{360DVD$^+$. Residual misalignment: faint blur bands around $\pm180^\circ$.}
\end{subfigure}\hspace{2pt}
\begin{subfigure}[t]{0.49\linewidth}
  \centering
  \includegraphics[width=\linewidth]{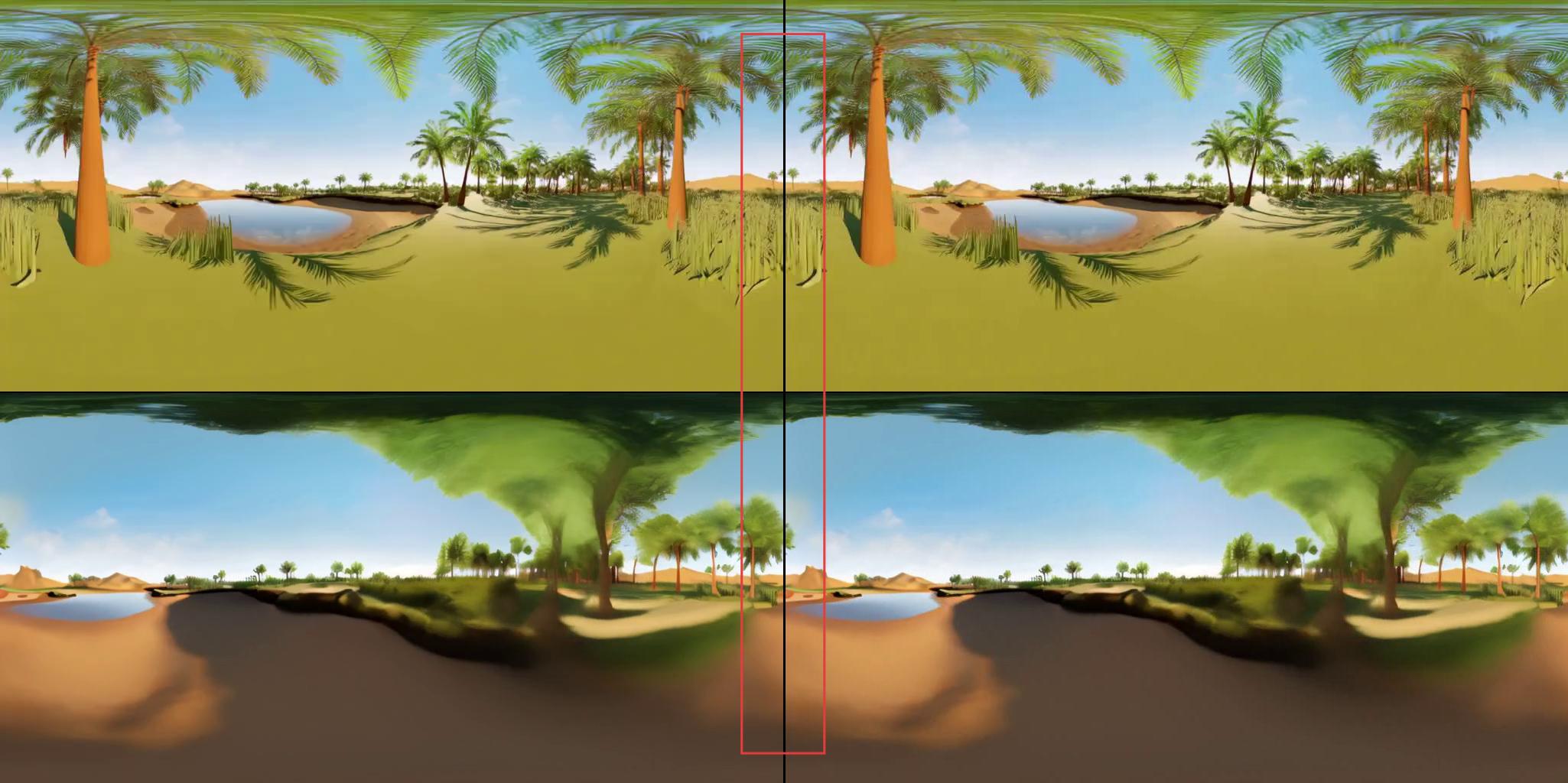}
  \caption{GenEX. Seam closure is decent, but textures are soft and details are lost.}
\end{subfigure}

\vspace{1pt} 

\begin{subfigure}[t]{0.49\linewidth}
  \centering
  \includegraphics[width=\linewidth]{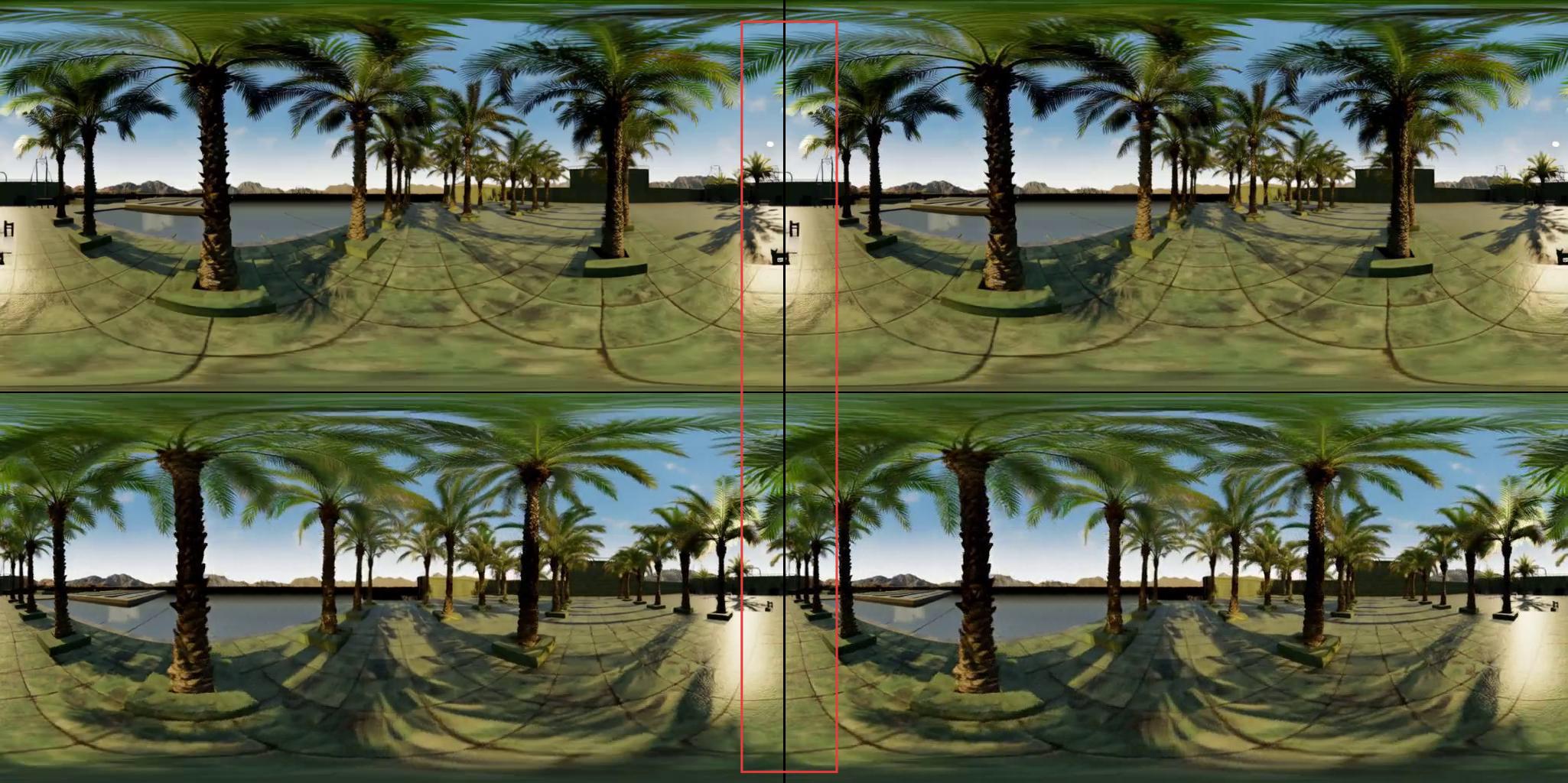}
  \caption{Ours. Clean seam, preserved high-frequency detail.}
\end{subfigure}\hspace{2pt}
\begin{subfigure}[t]{0.49\linewidth}
  \centering
  \includegraphics[width=\linewidth]{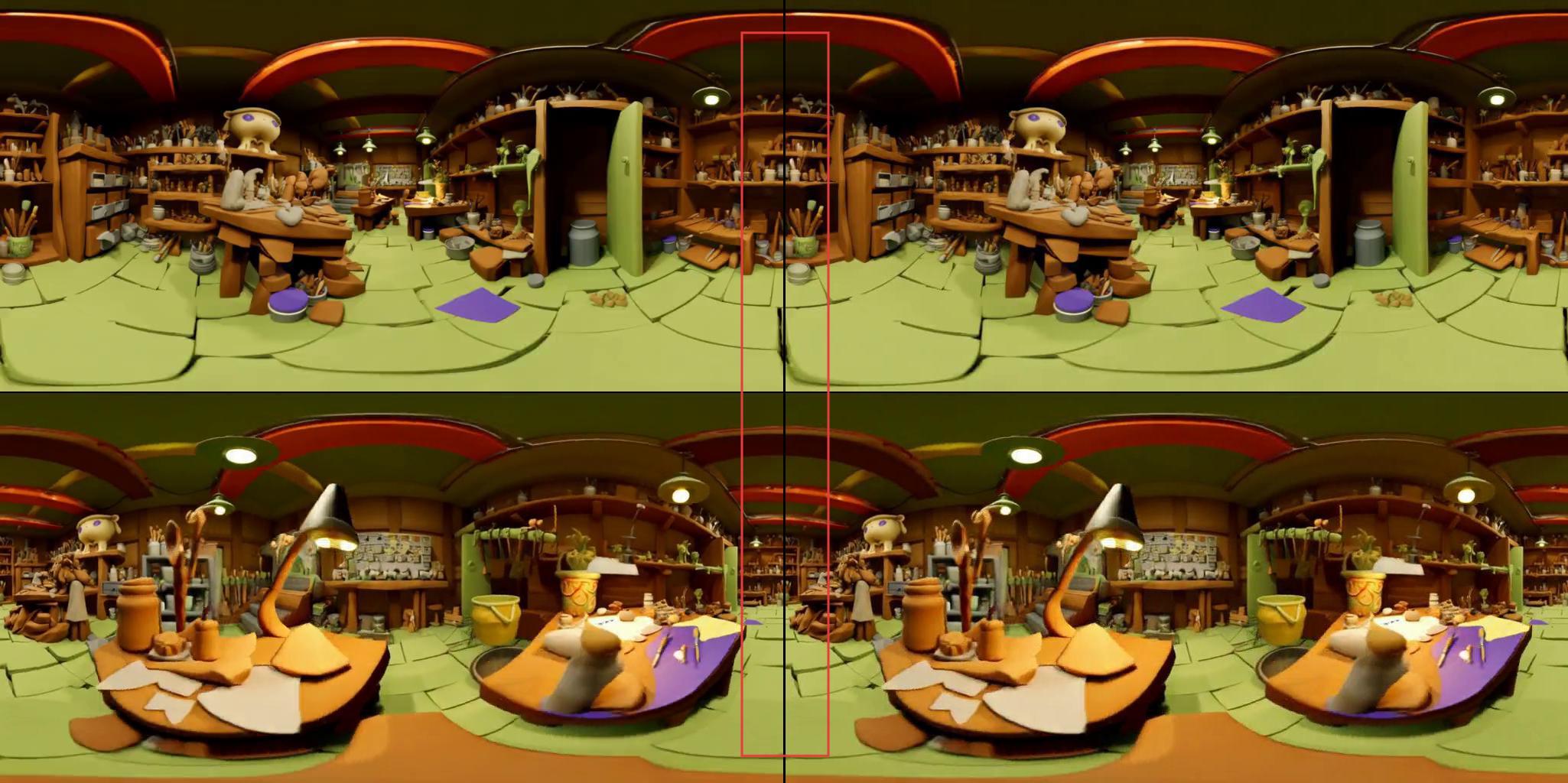}
  \caption{Ours (another scene). Consistent L--R alignment across frames.}
\end{subfigure}

\vspace{0.5cm}
\caption{Left--right boundary inspection over three consecutive frames. Our method achieves reliable seam closure while preserving fine details; 360DVD$^+$ and GenEX either show residual misalignment or detail loss.}
\label{fig:lr_alignment}
\end{figure*}

For temporal quality, motion magnitude connects directly to what is visible in the frames.
\textbf{360DVD} is nearly static across all directions ($<\!1.5$), which corresponds in Figures~\ref{fig:frames_compare} and \ref{fig:perspective} to \emph{minimal inter-frame displacement}: foreground trunks and the central path barely shift relative to the background, and near objects do not exhibit parallax.
\textbf{360DVD$^+$} raises motion slightly (e.g., front/back $\approx$1.2–1.5) and you can observe a faint forward drift, yet foreground elements still look \emph{pinned} in place; at the same time, Figure~\ref{fig:lr_alignment}(a) reveals faint blur bands at $\pm180^\circ$, consistent with its lower L--R consistency.
By contrast, \textbf{ours} produces the strongest motion across all views—front 4.02, back 3.99, left 3.56, right 5.11—approaching the GT range (6–8). 
This manifests in Figures~\ref{fig:frames_compare}(d–e) and \ref{fig:perspective}(b) as: 
(i) \emph{forward expansion} of the path and receding background in the front/back crops, 
(ii) \emph{lateral sweep} of trunks and water/ground edges in the left/right crops, and 
(iii) \emph{entry/exit of foliage} from the borders across time—hallmarks of depth-aware, camera-driven parallax.
Figure~\ref{fig:lr_alignment}(c–d) further shows that these larger motions are achieved while maintaining strict left–right alignment.

As a strong general baseline, \textbf{GenEX} matches our seam closure (L--R Cons.\,$=0.99$) but underperforms on motion, especially in the front/back views (1.91/1.48 vs.\ our 4.02/3.99). 
In Figure~\ref{fig:frames_compare}(c), its relatively higher right-view value (4.14) appears mainly as \emph{localized texture shimmer/wobble} rather than scene-scale parallax: foliage blurs and surfaces smear while the global depth layout barely progresses. 
This discrepancy explains why GenEX’s motion magnitude is not accompanied by coherent camera advancement.

\noindent\textbf{Qualitative comparison.}
Figures~\ref{fig:frames_compare} and \ref{fig:perspective} present visual evidence. For \textbf{360DVD} and \textbf{360DVD$^+$}, 360DVD produces semantically recognizable content but remains largely static (foreground trunks and central paths barely move; cyclic repetition appears). 360DVD$^+$ introduces limited motion yet still shows frozen foregrounds and faint seam blur near $\pm180^\circ$, consistent with its lower L--R consistency and small motion magnitude.
In contrast, \textbf{our method} (rank 16) generates sharper frames with coherent forward motion: trunks glide laterally, fresh foliage enters from the sides, and sunlight patches sweep smoothly across the ground. All four perspectives stitch into a \textbf{seamless panorama}, with objects crossing the $\pm180^\circ$ seam without tearing or flicker. The stronger parallax and fluid viewpoint transitions yield an immersive, 3D-aware experience, most evident in the challenging back-view trajectory.

Compared to \textbf{GenEX}, the gap is more pronounced. GenEX often produces plausible first frames but its quality degrades over time: palm trees and structures distort, foreground objects float or deform, and temporal coherence collapses. Visual clarity drops sharply—textures blur, details vanish, surfaces look overly smooth. While seam closure is acceptable, Figure~\ref{fig:lr_alignment} reveals that GenEX favors low-frequency structural alignment at the expense of high-frequency detail: foliage, terrain, and architectural textures appear flat or muddled. Our method maintains both global geometric consistency and fine-grained sharpness (e.g., clear, continuous boundaries around tree trunks and columns).

\paragraph{Conclusion.}
360DVD excels at static frame quality but lacks dynamic realism. 360DVD$^+$ benefits from LoRA injection yet remains constrained by the static-oriented design. GenEX is strong for general video synthesis but struggles with panoramic motion and fine detail preservation. In contrast, our LoRA-based adaptation achieves superior temporal coherence, richer inter-frame motion, and seamless panoramic stitching—improving both quantitative metrics and perceptual fidelity—thereby better serving real-world $360^\circ$ video generation.

\subsection{Ablation Study}

\paragraph{Ablation Study about LoRA Placement.}

To further dissect the contribution of different Transformer sub‑modules, we conduct an ablation study at a fixed rank of~16 and evaluate three variants: 
\begin{itemize} 
\item \textbf{Full} – LoRA adapters inserted in both attention and feed‑forward (linear) layers. 
\item \textbf{Attn‑only} – adapters applied exclusively to self‑attention projection matrices. 
\item \textbf{Lin‑only} – adapters applied only to MLP/linear layers. 
\end{itemize}
Besides 16, we also compare different numbers (5, 8, and 32) of the LoRA rank on panoramic video generation. For each setting, we report quantitative and qualitative results.

\begin{table}[h]
\centering
\caption{Ablations on LoRA placement (\textbf{w/o Linear}=Attn-only, \textbf{w/o Attention}=Lin-only) and capacity (rank). Metrics: left–right consistency (L–R Cons.) ($\uparrow$) and motion magnitude per view ($\uparrow$). \textbf{Ranks $<\!8$ show systematic degradation.}}
\label{tab:ablation_lora_parts}
\begin{tabular}{lccccc}
\toprule
\textbf{Variant} & \textbf{L--R Cons.} & \textbf{Front} & \textbf{Back} & \textbf{Left} & \textbf{Right} \\
\midrule
GT             & 1.00 & 7.64 & 6.70 & 7.71 & 7.69 \\
Attn-only  & 0.99 & 3.70 & 3.24 & \textbf{4.12} & 3.59 \\
Lin-only  & 0.99 & 3.62 & 3.06 & 3.60 & {4.48} \\
Full & \textbf{0.99} & \textbf{4.02} & {3.99} & 3.56 & {5.11} \\
Rank 5         & 0.74 & 1.32 & 3.73 & 1.58 & 3.20 \\
Rank 8         & 0.95 & 3.91 & 3.88 & 3.64 & 4.37 \\
Rank 32        & 0.90 & 3.92 & \textbf{5.00} & 3.41 & \textbf{5.88} \\
\bottomrule
\end{tabular}
\end{table}

\noindent\textbf{w/o Attention (Lin-only).}
Removing LoRA from attention and keeping it only in MLP/linear layers preserves seam closure (L--R Cons.\,=\,0.99) but skews the motion profile:
Front/Back/Left/Right = 3.62/3.06/3.60/\textbf{4.48} (Table~\ref{tab:ablation_lora_parts}).
Qualitatively, this appears as stronger right-view displacement but weaker forward/backward progression, suggesting that linear-only adapters emphasize appearance/texture modulation over true camera/parallax motion.
In the frames, lateral edges (e.g., water/ground boundaries) shift more than the path expansion in front/back, while seams remain clean.\\

\noindent\textbf{w/o Linear (Attn-only).}
Removing LoRA from linear layers and keeping it only on attention projections likewise maintains near-perfect seams (L--R\,=\,0.99) yet biases motion in the orthogonal direction:
Front/Back/Left/Right = 3.70/3.24/\textbf{4.12}/3.59.
Visually, the left-view lateral sweep is more apparent, but front/back trajectories advance more slowly than the default Full model.
This indicates that attention-only adapters mainly aid rotational/lateral alignment rather than depth-consistent forward motion.
Across examples, the $\pm180^\circ$ boundary stays closed, but parallax cues in front/back are reduced compared to the Full setting.\\
\begin{figure}[h]
\centering
\begin{subfigure}[t]{0.24\linewidth}
  \centering
  \includegraphics[width=\linewidth, height=5cm]{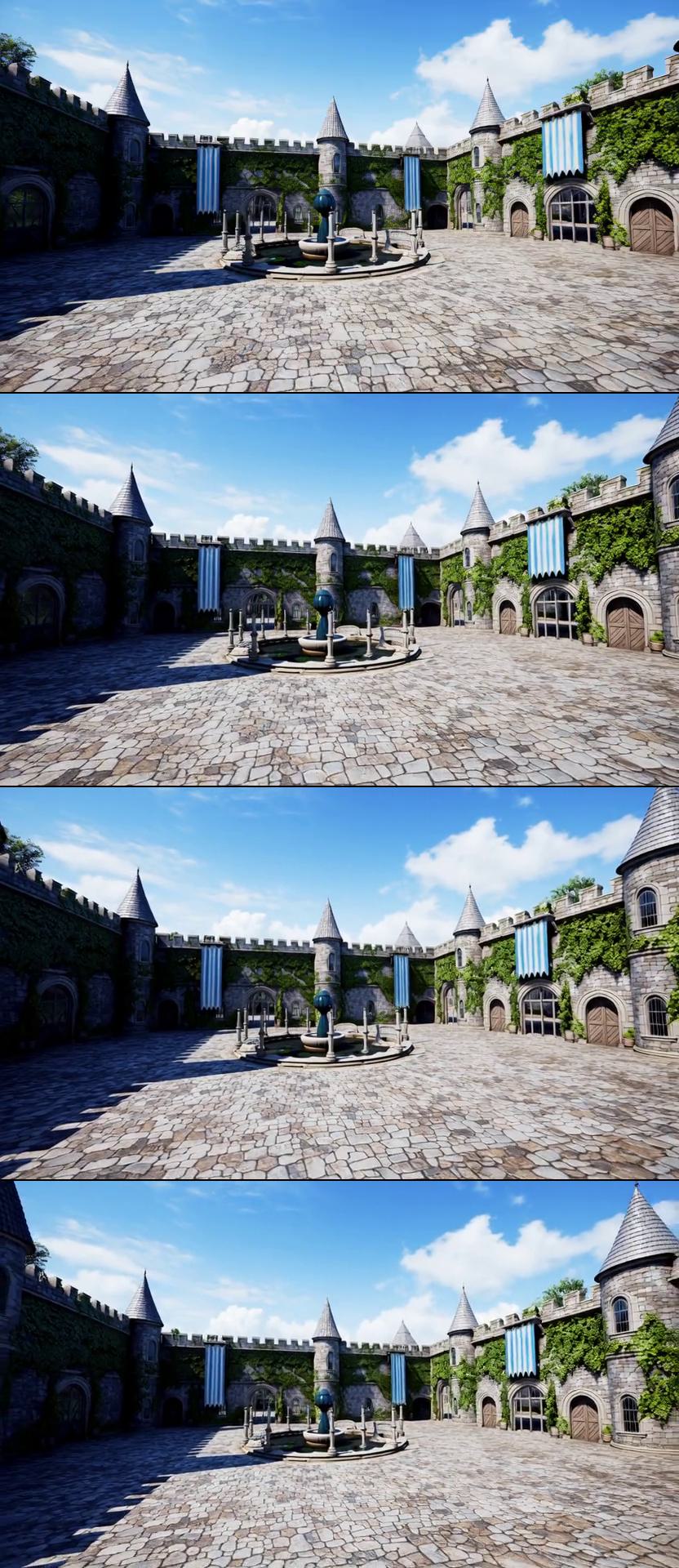}
  \caption{Rank 5}
  \label{fig:rank5}
\end{subfigure}
\begin{subfigure}[t]{0.24\linewidth}
  \centering
  \includegraphics[width=\linewidth, height=5cm]{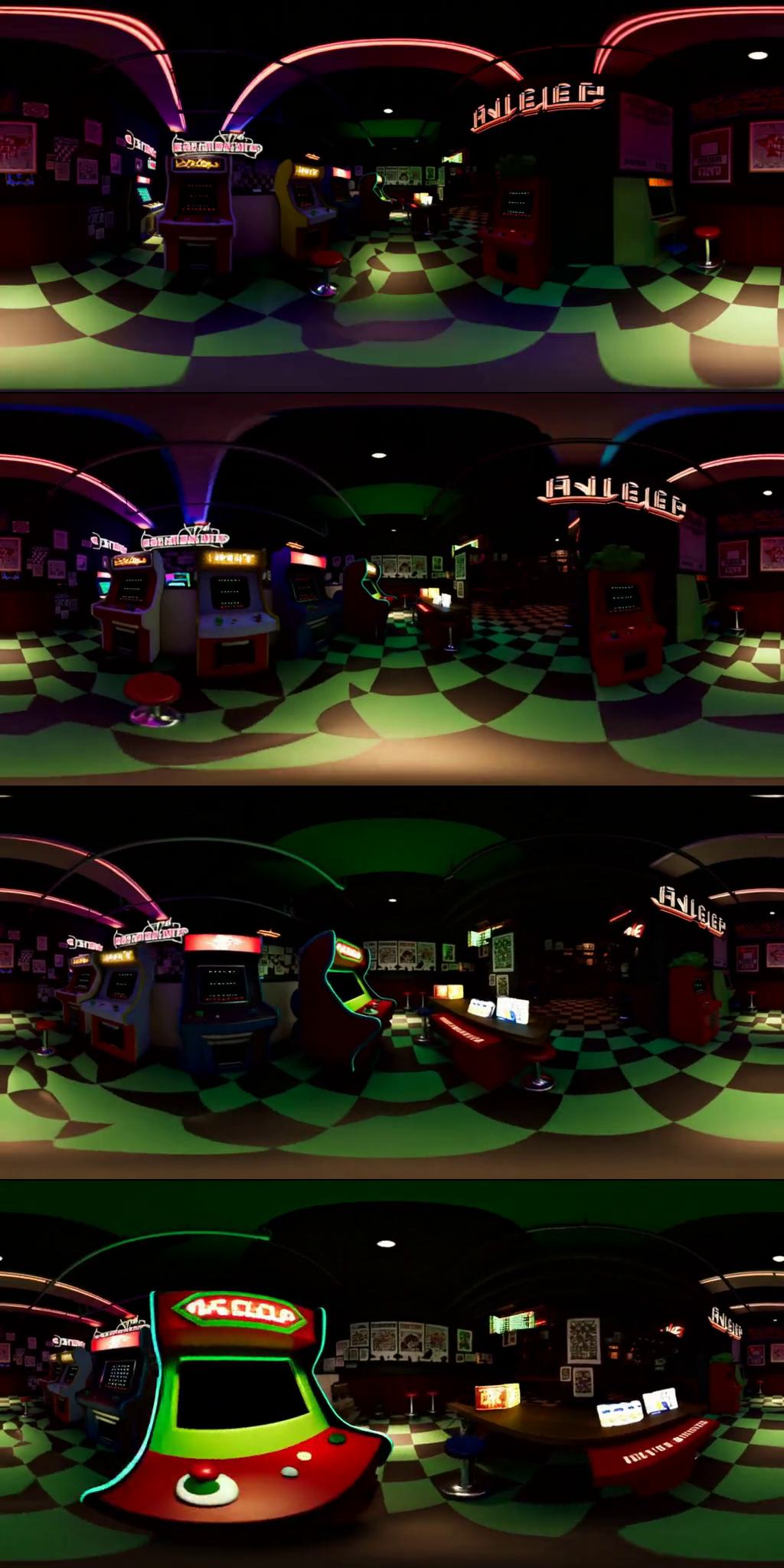}
  \caption{Rank 8}
  \label{fig:rank8}
\end{subfigure}
\begin{subfigure}[t]{0.24\linewidth}
  \centering
  \includegraphics[width=\linewidth, height=5cm]{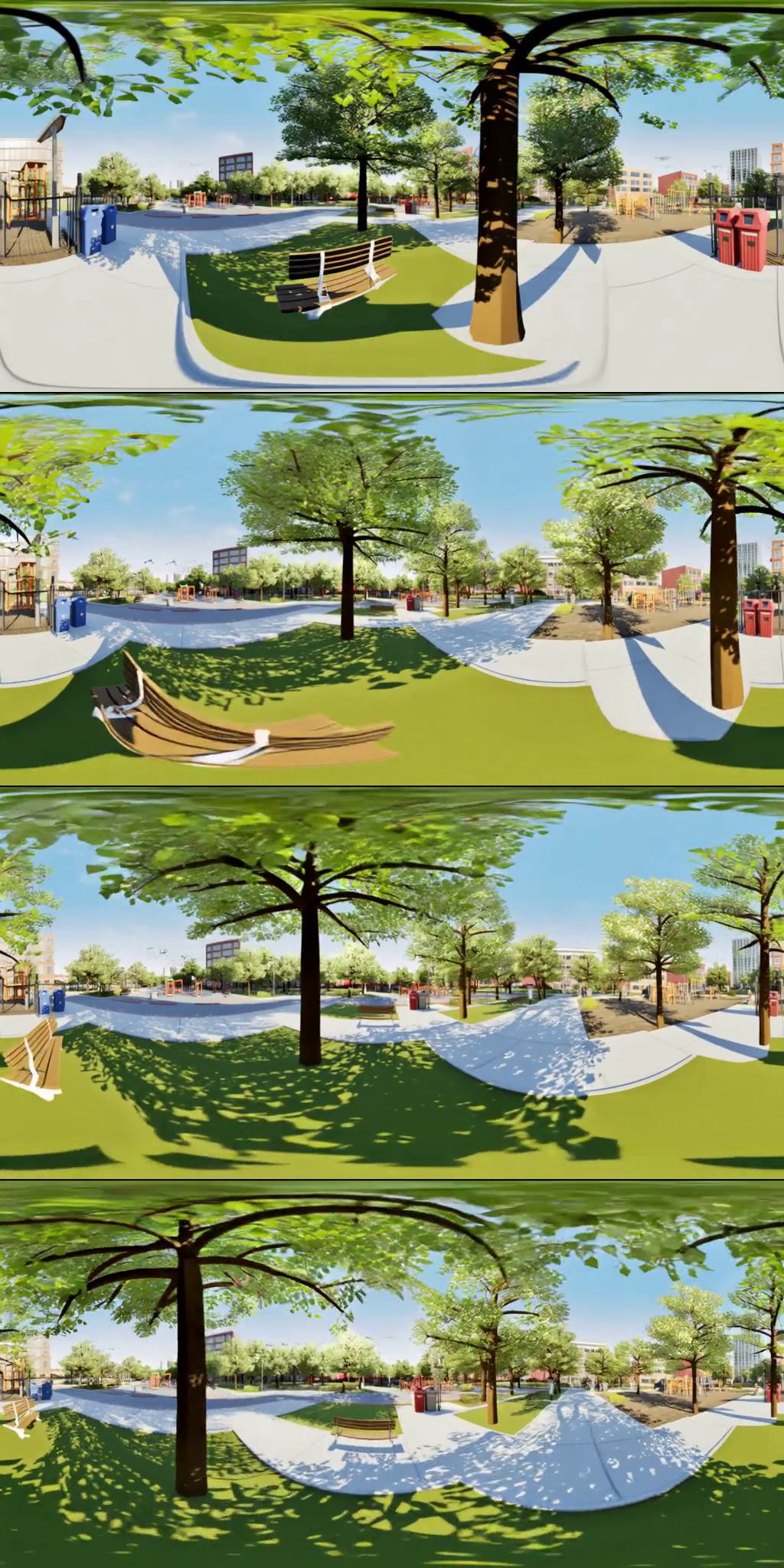}
  \caption{Rank 16}
  \label{fig:rank16}
\end{subfigure}
\begin{subfigure}[t]{0.24\linewidth}
  \centering
  \includegraphics[width=\linewidth, height=5cm]{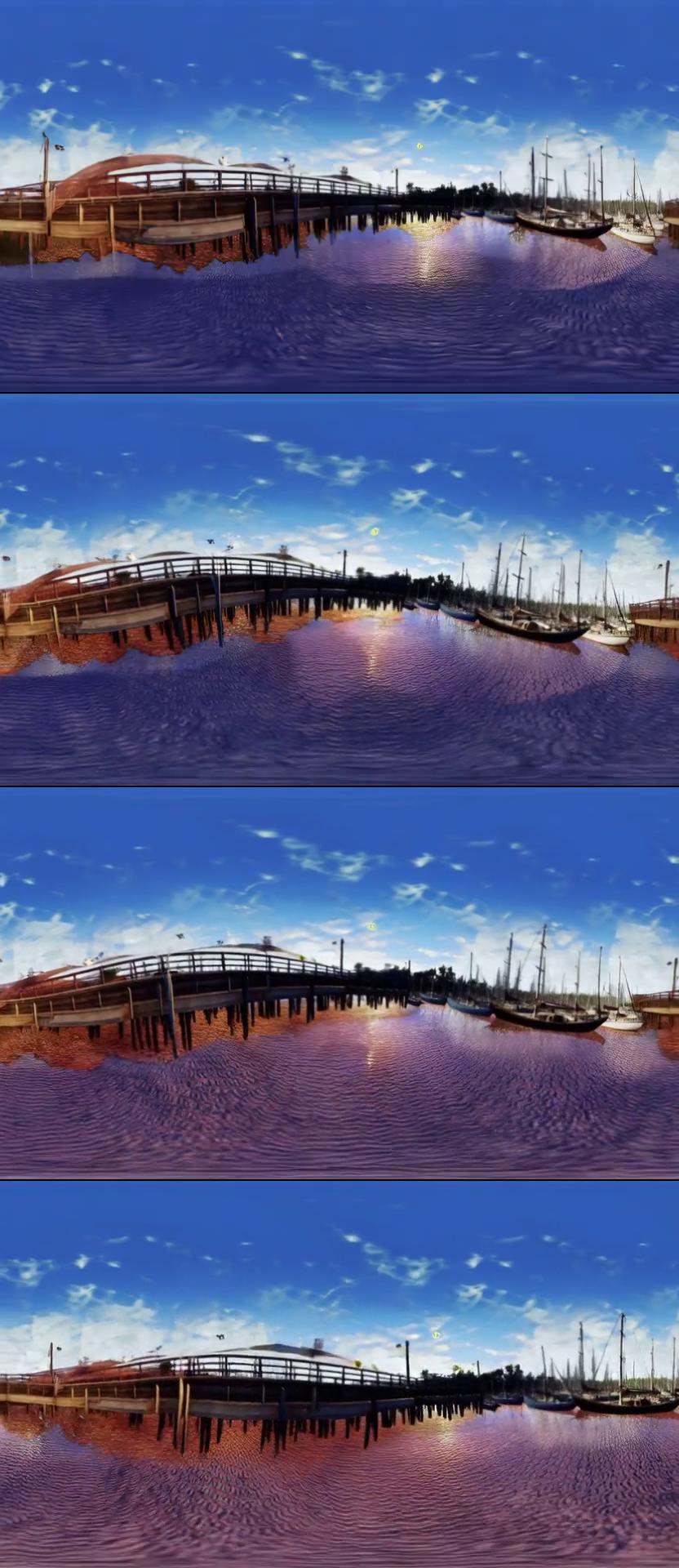}
  \caption{Rank 32}
  \label{fig:rank32}
\end{subfigure}
\caption{\textbf{Rank capacity threshold.} Ranks $<\!8$ underfit: seam breaks/repetitions and weak parallax (rank\,5 worst; rank\,8 partial recovery); rank\,16 is most stable; rank\,32 increases motion but introduces drift/flicker.}
\label{fig:different_rank}
\end{figure}
\paragraph{Ablation Study on Rank Tuning.} We observe that
ranks $<\!8$ fail to produce geometrically valid panoramas. At rank\,5, L--R Cons.\ drops to \textbf{0.74} with unstable motion (1.32/3.73/1.58/3.20), and the equirectangular wrap cannot close (visible discontinuities at $\pm180^\circ$);
\textbf{Rank\,8} partially recovers (L--R\,=\,0.95; 3.91/3.88/3.64/4.37) but still underperforms the default.
In contrast, the \textbf{Full (rank\,16)} configuration yields the best balance (L--R\,=\textbf{0.99}; \textbf{4.02}/\textbf{3.99}/3.56/\textbf{5.11}) with coherent forward motion and sharp textures.
Pushing capacity to \textbf{rank\,32} increases motion (Back=\textbf{5.00}, Right=\textbf{5.88}) but reduces alignment (L--R=0.90), manifesting as semantic drift and temporal flicker.
Overall, \textbf{rank\,16} is the most stable choice; \textbf{ranks $<\!8$} are underfit and fail to maintain global panoramic continuity.

\begin{figure}[H]
\centering
\begin{subfigure}[b]{\linewidth}
  \centering
  \includegraphics[width=\linewidth]{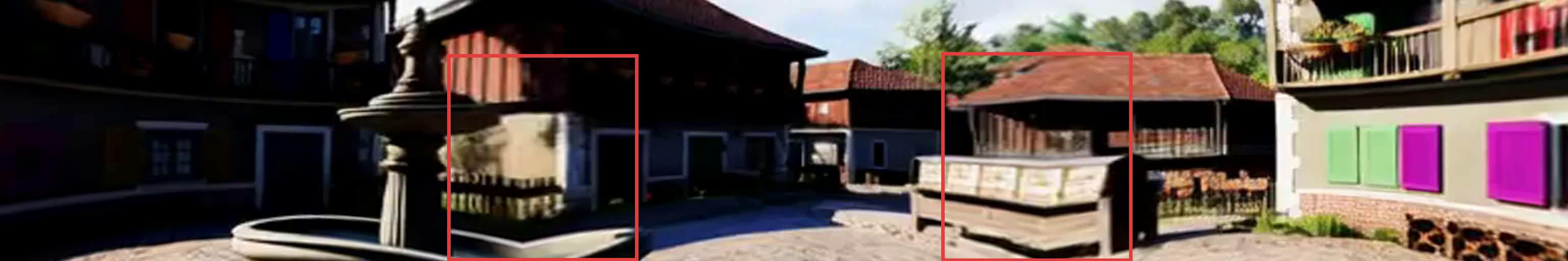}
  \caption{Rank 5}
  \label{fig:persp_5}
\end{subfigure}
\vspace{0.3em}
\begin{subfigure}[b]{\linewidth}
  \centering
  \includegraphics[width=\linewidth]{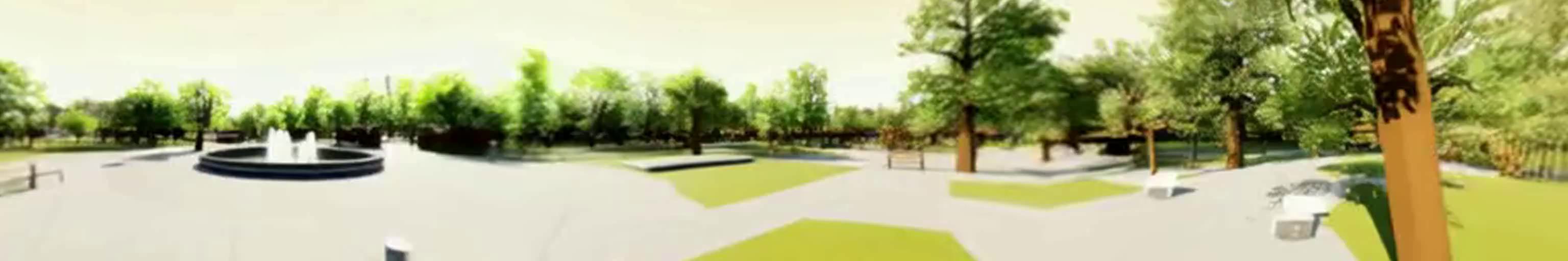}
  \caption{Rank 8}
  \label{fig:persp_8}
\end{subfigure}
\vspace{0.3em}
\begin{subfigure}[b]{\linewidth}
  \centering
  \includegraphics[width=\linewidth]{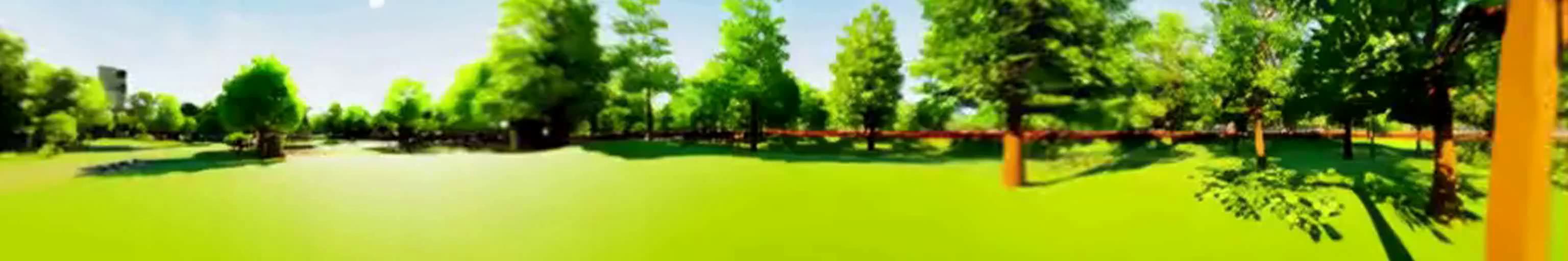}
  \caption{Rank 16}
  \label{fig:persp_16}
\end{subfigure}
\vspace{0.3em}
\begin{subfigure}[b]{\linewidth}
  \centering
  \includegraphics[width=\linewidth]{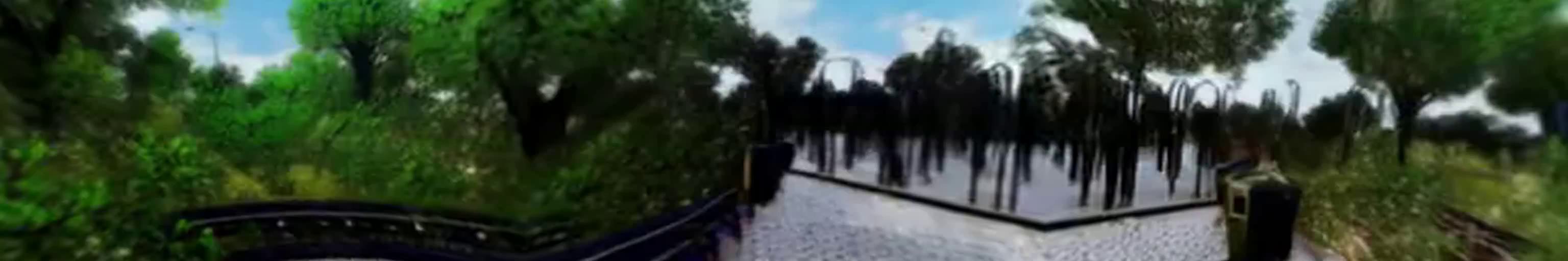}
  \caption{Rank 32}
  \label{fig:persp_32}  
\end{subfigure}

\vspace{0.5cm}
\caption{\textbf{Ranks $<\!8$ underfit}: at rank\,5 the panorama breaks down (blurred/flat textures, weak parallax); rank\,8 partially recovers but remains soft with limited depth cues.
\textbf{Rank\,16} delivers the best balance—sharp structures and stable parallax.
Pushing to \textbf{rank\,32} increases motion capacity but introduces drift/over-smoothness and geometric wobble.}
\label{fig:rank_perspective}
\end{figure}

\paragraph{Ablation summary.}
Removing either branch alone preserves seam closure (L--R $\approx$ 0.99) but \emph{biases} the motion profile: \textbf{w/o Attention} (Lin-only) favors appearance/texture modulation with weaker front/back progression, whereas \textbf{w/o Linear} (Attn-only) strengthens lateral/rotational alignment but reduces depth-consistent motion (Table~\ref{tab:ablation_lora_parts}). Using \textbf{both} components restores a balanced, 3D-aware trajectory with coherent parallax. Across ranks, a clear \textbf{capacity threshold} emerges: \textbf{ranks $<\!8$} underfit (down to L--R=\textbf{0.74}), causing seam breaks and weak/unstable parallax (Figure~\ref{fig:different_rank}); \textbf{rank\,16} offers the best trade-off (L--R=\textbf{0.99}; strong, balanced motion), while \textbf{rank\,32} increases motion but harms alignment (L--R=0.90) and induces drift/flicker. We therefore adopt the \textbf{Full} placement at \textbf{rank\,16} as the default for robust panoramic video generation.

\begin{figure}[h]
\centering
\begin{subfigure}[t]{\linewidth}
  \centering
  \includegraphics[width=\linewidth,height=0.18\textheight]{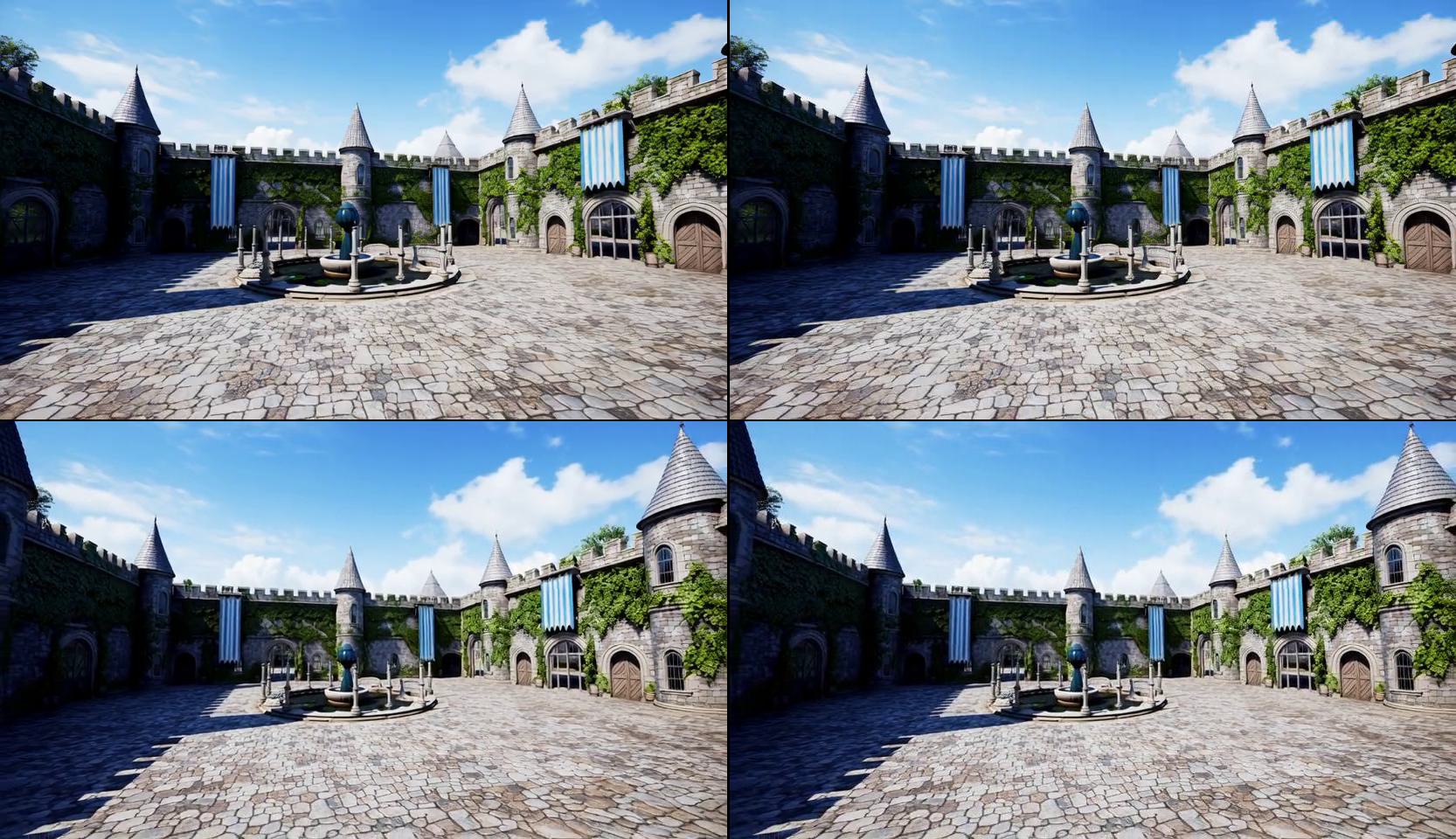}
  \caption{Rank 5 — seam \emph{not} closed; misalignment at $\pm180^\circ$ (L--R Cons.\ 0.74).}
\end{subfigure}
\vspace{0.5cm}
\begin{subfigure}[t]{\linewidth}
  \centering
  \includegraphics[width=\linewidth,height=0.18\textheight]{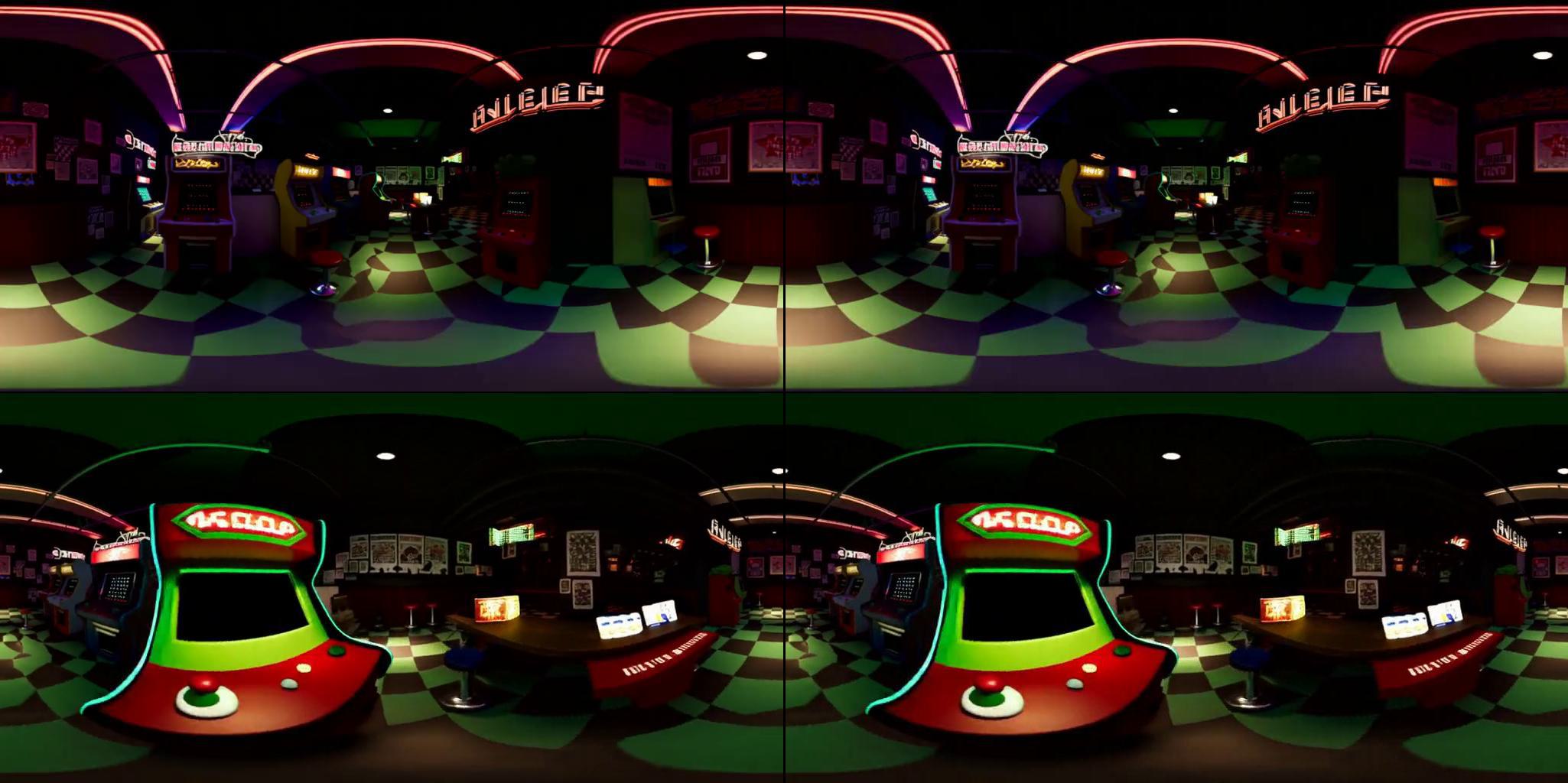}
  \caption{Rank 8 — left–right seam aligned; panorama closes (L--R Cons.\ 0.95).}
\end{subfigure}
\vspace{0.5cm}
\caption{\textbf{Effect of rank on left–right seam.}
Lower capacity (rank 5) cannot form a valid panoramic wrap, causing breaks at the $\,\pm180^\circ$ boundary; increasing to rank 8 restores left–right alignment and a closed seam.}
\end{figure}

\section{Conclusion}
We propose a novel approach to panoramic video generation by reformulating it as a lightweight LoRA-based adaptation task, enabling efficient fine-tuning of pretrained video diffusion models. Our theoretical analysis confirms that LoRA can effectively model the perspective-to-panoramic transformation when its rank exceeds the task’s inherent degrees of freedom. Experiments demonstrate that our method achieves high-quality panoramic video synthesis with minimal training data, outperforming prior works in visual fidelity, consistency, and motion diversity. This work bridges the gap between standard and panoramic video generation, advancing immersive 3D content creation.


\newpage
\bibliography{main}

\end{document}